\documentclass[11pt]{article}

\usepackage[utf8]{inputenc}
\usepackage[T1]{fontenc}
\usepackage{lmodern}
\usepackage{microtype}
\usepackage{graphicx}
\usepackage{float}
\usepackage{booktabs}
\usepackage{array}
\usepackage{amsmath}
\usepackage{url}
\usepackage[hidelinks]{hyperref}
\usepackage[numbers,sort&compress]{natbib}
\usepackage[margin=1.15in]{geometry}

\title{Search Discipline for Long-Horizon Research Agents}

\author{
\begin{tabular}{cc}
Adithya Srinivasan$^*$ & Devesh Paragiri$^*$\\
\small Department of Computer Science & \small Department of Computer Science\\
\small North Carolina State University & \small University of Maryland\\
\small \texttt{asrini27@ncsu.edu} & \small \texttt{deveshp@umd.edu}\\[1.4em]
\multicolumn{2}{c}{\small $^*$Equal contribution.}
\end{tabular}
}
\date{}

\begin{document}
\maketitle

\begin{abstract}
Autoresearch agents now propose, evaluate, and select scientific candidates against a metric, and that metric is usually an aggregate reduced over a heterogeneous space of regions, slices, or cohorts. We show that when scientific validity lives in that disaggregated structure, the aggregate can rank the wrong candidate first. The headline number improves while the structure underneath inverts, so a decision made on the number accepts a candidate that quietly breaks the model. The failure is not domain-specific. It appears wherever a candidate's validity is multi-dimensional but its verifier is a single reduction.

We demonstrate the inversion on a fire-model task in the Ecosystem Demography model. The highest-scoring candidate and a slightly lower one are within noise of each other on global score, yet the top-scoring one collapses the protected boreal regions while the other preserves them. What separates them is the per-region behavior, not the headline number.

This decision should not be left to the agent that produced the candidates. The agent optimizing the score is the last party likely to catch the score being wrong, and a prompt has no remaining turn once the agent has stopped. We move the decision to an external control loop that audits each candidate on its disaggregated behavior and acts after the agent has decided. It can demote a candidate the agent would have accepted, and it can reopen a run the agent had declared finished. Our contribution is the inversion finding itself, and a search-discipline protocol that decides on reviewable candidate-effect evidence instead of the score.
\end{abstract}

\section{Introduction}

Autoresearch agents now run long loops. They reproduce a baseline, edit a model, search over its parameters, evaluate the result, and write up what they found. The hard part is the decision that comes after a candidate improves the score the agent is optimizing. The same improvement can mean a real mechanism, a fix that helps the average while breaking an important part of the input space, an exploit of the evaluation, or a narrow repair that misses the failure the search was supposed to address.

This decision is hard because the score the agent optimizes is usually an aggregate, a single number reduced over a heterogeneous space of regions, slices, or cohorts. When the scientific validity of a candidate lives in that disaggregated structure, the aggregate and the correct decision can come apart. The headline number improves while the structure underneath gets worse, and a choice made on the number selects a candidate that quietly breaks the model. We call this an inversion, and it is the central object of the paper. It can arise whenever a candidate has to be right along several dimensions at once while the verifier reports only one.

Inside the agent, the decision sits in the weakest possible place. The agent that produced the candidate also judges it, in its own context and on its own schedule, while optimizing the very score that has inverted. A prompt does not fix this. A prompt is consumed at the start of the run; it has no way to wake when an evaluation finishes and no remaining turn once the agent has stopped. The party optimizing the score has the weakest position from which to notice that the score is wrong.

We therefore place the decision outside the agent, in a control loop that audits each serious candidate on its disaggregated behavior and acts at points the agent does not control. We call the target behavior search discipline, and the rest of the paper develops the audit it enforces and the loop that carries it.

This work is closest to long-horizon agent evaluation and to scientific-agent systems. SWE-bench and MLE-bench measure whether agents complete realistic software and machine-learning tasks \citep{swebench,mlebench}. Scientific-agent systems and case studies show agents generating experiments, code, and written reports under human or automated validation \citep{aiscientistv2,alphaevolve,nguyen2026physics}. Our focus is narrower and earlier. It is the moment a candidate with mixed effects has to be accepted, demoted, or sent back for more search.

This paper does three things. It names and demonstrates aggregate-verifier inversion on a real scientific-modeling task, in a run where the top-scoring candidate is the one that breaks the model. It proposes search discipline, deciding on a candidate-effect audit in place of the score. And it builds the control loop that enforces this from outside the agent, which is what lets the run override an accept decision the agent had made and reopen one it had stopped, the two moves a prompt cannot make.

\section{The inversion problem}

We treat an autoresearch run as a search for a candidate that improves a target system under a fixed contract. The contract fixes what the candidate may use, what it may not, how it is evaluated, and when the search may stop. The agent proposes candidates, evaluates them, and selects one.

The candidate is scored by a verifier. In the cases we care about the verifier is a reduction that collapses behavior over a heterogeneous space into one aggregate number. We distinguish the aggregate score, the single number the agent optimizes, from the disaggregated behavior, what the candidate does across the parts of the space, the regions, slices, or cohorts it actually has to get right. Scientific validity is a judgment on the disaggregated behavior, not on the aggregate.

An inversion occurs when the aggregate and the validity judgment disagree on which of two candidates is better. The aggregate prefers the higher-scoring candidate, while the other is the defensible one, because the higher-scoring candidate buys its score by damaging a part of the space that matters. The aggregate is not merely noisy here. It points the wrong way. Any selector that reads only the aggregate, whether the agent, a prompt, or an external check that gates on the same number, takes the candidate that breaks the model.

Two properties make this likely in autoresearch rather than rare. First, the space is heterogeneous, and a candidate can help easy parts and harm hard ones, and a mean over both can rise while the hard parts collapse; the same hidden-stratification effect makes an aggregate metric miss subgroup failures in standard evaluation \citep{hiddenstratification}. Second, the aggregate has limited resolution. Near a strong baseline, candidate-to-candidate differences in the aggregate are often within the noise of the evaluation, reachable by parameter tuning alone, so the aggregate stops carrying usable signal exactly when the disaggregated differences become decisive. When both hold, the decision cannot be read off the score, and it has to be made on the disaggregated behavior.

\section{Search discipline and the control loop}

Search discipline means deciding on disaggregated behavior and keeping the run honest about what it has and has not ruled out. It has two parts, an audit that defines the evidence a decision must rest on, and an external loop that enforces the audit at the points where the decision is actually made.

\subsection{The candidate-effect audit}

For each serious candidate the run must produce a structured account in place of a score. The account records where the candidate helps and where it harms across the space, the unchanged parts, the movement in each component of the verifier, the legal input-derived state that separates the helped parts from the harmed parts, and a bounded diagnostic that tests that separator. The account resolves into one of five roles: (1)~a \emph{score winner} that tops the aggregate but fails the disaggregated test; (2)~a \emph{defended candidate} that holds up on both; (3)~a \emph{tradeoff} that repairs one part of the space while damaging another; (4)~an \emph{informative failure} that rules a direction out; or (5)~a \emph{rejected shortcut}. The run is accountable to the role, and a score alone does not satisfy it.

\subsection{Why the decision must be external}

The audit is a rubric, and a rubric can be written into a prompt. What a prompt cannot supply is the position from which the audit is applied. In-context self-critique and iterative self-refinement keep the evaluation inside the agent that produced the candidate \citep{reflexion,selfrefine}. That agent evaluates its own work while optimizing the aggregate that has inverted, so its self-assessment shares the bias that created the problem. A prompt also has no turn after the agent stops. Once the agent writes its result and ends, the prompt cannot reopen the run. Placing the audit in a separate process makes the judge a party that did not generate the candidate and has no stake in it, and gives it the ability to act after the agent has already decided. These are properties of where the decision sits, and better phrasing of a prompt does not supply them.

\subsection{The control loop}

\begin{figure}[H]
  \centering
  \includegraphics[width=0.94\columnwidth]{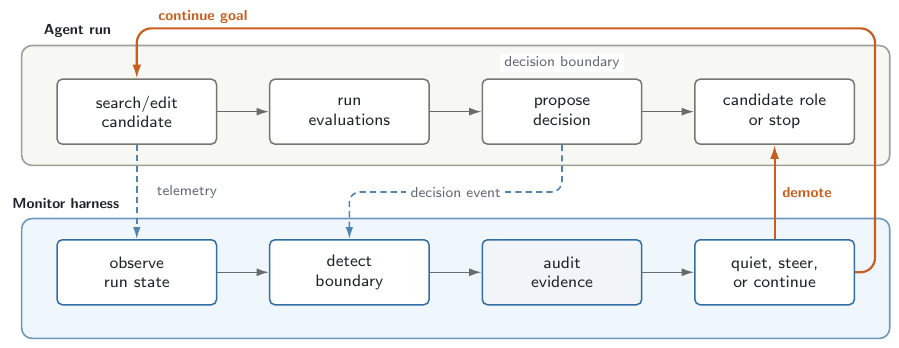}
  \caption{The monitor observes the agent run and acts at decision boundaries. The orange arrows mark the two actions a one-time prompt cannot take after the agent has already decided, demoting a candidate and reopening the run with a continuation.}
  \label{fig:loop}
\end{figure}

The loop wraps an otherwise normal agent run. The agent does the research. The monitor observes the run, detects when the agent has reached a decision point, sends a compact summary of the evidence to a judge, and then stays quiet, sends a corrective steer, or sends a continuation that reopens the run. The judge is a separate language model used as an evaluator \citep{llmjudge}, and here it is advisory. It returns a structured decision, including whether the contract is satisfied, which candidate role applies, and whether to intervene, and the monitor renders and delivers any message, so a judge error cannot reach the agent unfiltered. The intervention is the whole loop, with the judge as one part. What the loop observes, when it decides to act, and how it delivers the action all determine what it can do.

\subsection{Making the loop reliable}

A few design choices are what keep the loop from disrupting the run it is supposed to govern. The monitor launches the agent itself and owns its terminal, so it has a reliable channel to deliver instructions and to continue or stop the run, going beyond only watching it. It judges at completed evidence, using process completion as the signal and requiring result files to be present and stable before it acts, so the judge sees finished work. It delivers long instructions by writing them to a file and pointing the agent at it, which avoids partial reads and leaves an artifact for audit. Finally, a run counts as finished only when the loop records a terminal status; an agent writing its own final report does not end it, and the most common stall is an agent that returns to idle believing it is done, which the loop catches and reopens if the contract is unmet. When a reopen collides with the agent's own shutdown, the run can be scientifically complete yet uncertified, and we keep those cases separate from cleanly certified ones.

\section{Evaluation}

\subsection{Task and setup}

We study the inversion on a scientific-modeling task, improving the fire component of a terrestrial ecosystem model, which predicts how much area burns, under a fixed input contract \citep{moorcroft2001ed,ma2022ed,gel_ed}. The task is a good test because validity is genuinely disaggregated. The model is scored both by an overall number and by its behavior across world regions, and a candidate can move the overall number while changing different regions in opposite directions. The agent starts from a strong baseline and searches for a better global formula using only a fixed set of model-derived inputs such as dryness, rainfall, temperature, and vegetation productivity. It may not write rules that single out named regions, use latitude or longitude directly, add outside data, build per-location lookup tables, or fit the target directly. Serious candidates are scored globally and per region, with an independent public comparison as a cross-check \citep{paragiri2026fire}.

The task contains the inversion by construction. The model over-predicts burning in several low-fire regions, so the obvious move is to add a suppressor that lowers predicted burning. The same suppressor can damage the regions that genuinely burn, such as boreal forest, where a lower prediction is wrong. A suppressor can therefore raise the overall score by fixing the easy low-fire regions while quietly breaking the regions that matter, and the overall score will not reveal it. The region groupings we use below are for analysis only; the agent never sees them, and cannot target them.

We compare three conditions. The first is a strong base prompt that already requires baseline reproduction, global and regional checks, a single legal formula, evaluation, and maintained logs. The second adds one structural-reframing paragraph to that prompt. The third runs the agent under the control loop. The base prompt is demanding by design, and agents under it find real mechanisms on their own, so the loop competes against a capable control.

Across all conditions the research agent is Hermes \citep{hermesagent}, an open-source autonomous agent, driven by GPT-5.5 with reasoning effort set to high \citep{gpt55}; the monitor's judge uses the same model.

\subsection{Observing the inversion}

In the run where we see the inversion most cleanly, the agent produced two serious candidates from the same mechanism family, one tuned more aggressively than the other. On the overall score the two are indistinguishable. They sit within $0.006$ of the baseline and within $0.001$ of each other. That gap is reachable by tuning alone and is too small to support a claim of improvement. The overall score gives no reason to prefer either candidate.

The regional behavior makes the choice obvious in the opposite direction. The higher-scoring candidate collapses the boreal forest regions and drops their fidelity by about $0.10$ and $0.07$. These are first-decimal failures in regions that are supposed to burn. The slightly lower-scoring candidate holds those same regions near the baseline, within about $0.006$ and $0.003$, and keeps comparable repairs elsewhere.

The two numbers live on different scales, and that is the whole point. The global score is a mean pooled over every region, so a real change in one region is diluted into a fourth-decimal movement; a $0.0007$ gap there is within noise. The boreal figure is that region's own score, undiluted, so a $0.08$ change is a structural failure well above the region's noise. The aggregate stays flat precisely because it averages the boreal collapse together with repairs elsewhere, which is what hides the damage (Table~\ref{tab:inversion}). An agent or a prompt selecting on the overall score takes the higher number and ships the candidate that breaks the boreal forest.

\begin{table}[t]
  \centering
  \small
  \setlength{\tabcolsep}{0pt}
  \renewcommand{\arraystretch}{1.22}
  \begin{tabular}{@{}l@{\hspace{2.6em}}l@{\hspace{2.6em}}l@{\hspace{2.6em}}l@{}}
    \toprule
    Candidate & Global score & Boreal change & Decision \\
    \midrule
    Aggressive & $0.6774$ (highest)   & $-0.086$ & \textsc{demote} \\
    Gentler    & $0.6767$ ($-0.0007$) & $-0.004$ & \textsc{accept} \\
    \bottomrule
  \end{tabular}
  \caption{The aggregate score picks the wrong candidate. The two candidates are separated by $0.0007$ on the global score, within noise, but by about $0.082$ on boreal-region fidelity, a first-decimal change in a single region. The candidate that wins the aggregate is the one that breaks the boreal forest, and the loop accepts the lower-scoring candidate that preserves it.}
  \label{tab:inversion}
\end{table}

Under the loop, the judge declined the higher-scoring candidate on regional grounds and the run accepted the lower-scoring, regionally sound one. A party outside the agent refused the score winner on the disaggregated evidence, which is the override a prompt cannot produce, and the decision is recorded in the run's logs. We present this as one clean instance of the inversion and of the loop acting on it, not as an averaged effect.

\subsection{Pushing past premature stopping}

A separate failure the loop addresses is stopping too early. Under the base and reframed prompts, the agent tends to settle on the first mechanism that looks defensible and write its report after a single shallow pass. The loop changes this where it is logged. In one run it rejected a finished-looking state, forced another pass through combinations the agent had not tried, and only then certified a conservative outcome that kept the baseline. In another it blocked a stop the agent attempted at a candidate whose score sat within noise of the baseline, a difference too small to justify ending the search, and required more work before the run could close. In both cases the search continued past the point the agent would have stopped, and the push came from outside the agent after it had already returned to idle. The first run still kept the baseline as its answer, which shows the loop forces more search without manufacturing a result.

\subsection{Candidate roles across runs}

The inversion is not a one-off. Across the runs we examined, the global score stays within noise of the baseline for almost every serious candidate, so the score never settles the choice and the audit has to. The audit recovers a small set of research states the score cannot tell apart: a score winner that tops the aggregate while failing the regional test, a defended candidate that survives both, a tradeoff that repairs one part of the space while damaging another, and an informative failure that rules a direction out. These are different outcomes, and they are invisible to the number.

We had expected a cheaper route to work. We first tried prompt-level structural reframing, a single paragraph that asks the agent to draw on mechanisms from other domains, on the hypothesis that a better search prior would push the frontier. It did not. Reframing moved the agent earlier toward composing several mechanisms together, but the run still stopped after one shallow pass and the final frontier was unchanged. A strong base prompt with no loop also found a real mechanism on its own, while still damaging the boreal regions. Prompting can change where a search starts, but the agent still signs off on its own candidate, and taking that final decision out of the agent's hands is what moved us to an external loop.

\subsection{When the metric cannot separate candidates}

Across all conditions, fixing the low-fire regions without damaging the regions that genuinely burn is hard under the fixed input contract. This is not a hard ceiling, since some candidates improve several parts of the space at once, but the allowed inputs separate some cases and not others. They can identify a dry-season burning window, for example, yet they do not contain every cause of regional fire behavior. Causes the inputs omit, such as human ignition and land use, lightning, wind, and vegetation structure, may be why a blunt suppressor cannot win cleanly. The audit makes this boundary visible. When no clean candidate exists, the run still records which separations were tried and why each failed, which is more useful to whoever picks up the work than a single number that stalled.

\section{Discussion}

The loop governs the decision; it does not produce the candidates. A strong base prompt already finds real mechanisms, and one of our strongest candidates came from a run with no loop. What the loop adds is two operations the producing agent cannot reliably perform on itself. It keeps the run searching past the first defensible result, so the search does not end at the first thing that looks good. And it moves the accept decision to a party that did not produce the candidate, so a candidate can be demoted on disaggregated grounds even when its aggregate is the best in the run. A candidate can be a legitimate mechanism and still be the wrong answer to ship, and the loop turns that judgment into an explicit, recorded step taken from outside the agent.

The same shape appears whenever an autoresearch loop optimizes an aggregate over a heterogeneous space. A model-improving agent can raise validation accuracy while degrading a protected subgroup or quietly exploiting leakage, and a coding agent can turn the visible tests green while breaking behavior no test covers. The aggregate the agent watches is the wrong place to catch either. The decision turns on the same disaggregated questions we ask here, namely what improved, what got worse, what legal factor explains the split, and what role the candidate plays.

Our scope is narrow. The clearest evidence is a single certified run in which the inversion appears and the loop acts on it, supported by further runs whose candidate-effect traces show the same reading even where the run was not cleanly certified. We present the inversion as a demonstrated phenomenon and the loop as a working way to act on it, and leave to future work how often the loop changes an outcome across many runs. The most informative evidence is where the audit changes a later decision, demoting a candidate or reopening a search, which the volume of logging alone does not explain. The contract itself is a real boundary. If the allowed inputs do not contain the missing causes of behavior, no audit can recover them, and the audit's job is then to expose that boundary. The audit surfacing it is the system working, not failing.

When an autoresearch agent optimizes an aggregate metric, the metric can rank the wrong candidate first, and the agent optimizing it is the last party to notice. Search discipline answers this by deciding on disaggregated evidence and by placing that decision outside the agent, where a control loop can refuse the highest-scoring candidate for a defensible one and reopen a search the agent had ended. The standard it enforces is a reviewable account of what was searched and why each candidate was kept or ruled out, with a high score treated as a claim to be checked rather than a decision. That standard applies to scientific agents, machine-learning agents, coding agents, and any agent whose final answer has to survive review.

\bibliographystyle{plainnat}
\bibliography{references}

\end{document}